\documentclass[10pt,twocolumn,letterpaper]{article}
\usepackage[toc,page]{appendix}
\usepackage{cvpr}              

\usepackage{bbm}
\usepackage{float}
\usepackage{pifont}
\usepackage{comment}
\usepackage{amsmath}
\usepackage{amssymb}
\usepackage{booktabs}
\usepackage{enumitem}
\usepackage{graphicx}
\usepackage{subcaption}

\newcommand{\xmark}{\ding{55}}%
\usepackage[pagebackref,breaklinks,colorlinks]{hyperref}

\usepackage[capitalize]{cleveref}
\crefname{section}{Sec.}{Secs.}
\Crefname{section}{Section}{Sections}
\Crefname{table}{Table}{Tables}
\crefname{table}{Tab.}{Tabs.}

\def\cvprPaperID{2926} 
\def\confName{CVPR}
\def\confYear{2022}

\begin{document}


\title{It's About Time: Analog Clock Reading in the Wild}

\author{Charig Yang$^1$ \qquad \quad Weidi Xie$^{1,2}$ \quad \qquad Andrew Zisserman$^1$ \\ [5pt]
$^1$VGG, Department of Engineering Science, University of Oxford 
$^2$Shanghai Jiao Tong University \\
{\tt\small \{charig,weidi,az\}@robots.ox.ac.uk}\\
 {\url{https://www.robots.ox.ac.uk/~vgg/research/time}}}

\maketitle

\begin{abstract}
In this paper, 
we present a framework for reading analog clocks in natural images or videos. 
Specifically, we make the following contributions:
{\em First}, we create a scalable pipeline for generating synthetic clocks, 
significantly reducing the requirements for the labour-intensive annotations;
{\em Second}, we introduce a clock recognition architecture based on spatial transformer networks~(STN),
which is trained end-to-end for clock alignment and recognition.
We show that the model trained on the proposed synthetic dataset generalises towards real clocks with good accuracy, advocating a Sim2Real training regime;
{\em Third}, to further reduce the gap between simulation and real data,
we leverage the special property of ``time'', {\em i.e.}~uniformity,
to generate reliable pseudo-labels on real unlabelled clock videos, 
and show that training on these videos offers further improvements while still requiring zero manual annotations.
{\em Lastly}, we introduce {\bf three} benchmark datasets based on COCO, Open Images, 
and The Clock movie,
with full annotations for time, accurate to the minute.
\end{abstract}

\section{Introduction}
\label{sec:intro}

Humans are able to sense the time to some level of granularity given environmental cues, such as luminance or the extent of shadows. 
However, in order to know the \textit{exact} time we read from a time-keeping instrument such as a clock or watch.
Clocks come in different shapes, forms and styles, 
and humans are able to read them despite not having seen the particular clock before. In this paper our objective is to  enable a machine to perform the same task of telling the time from clocks {\em in the wild}.

\begin{figure*}[t]
{
    \centering
    \vspace{5pt}
    \begin{subfigure}[b]{0.33\textwidth}
         \centering
         \includegraphics[width=\textwidth]{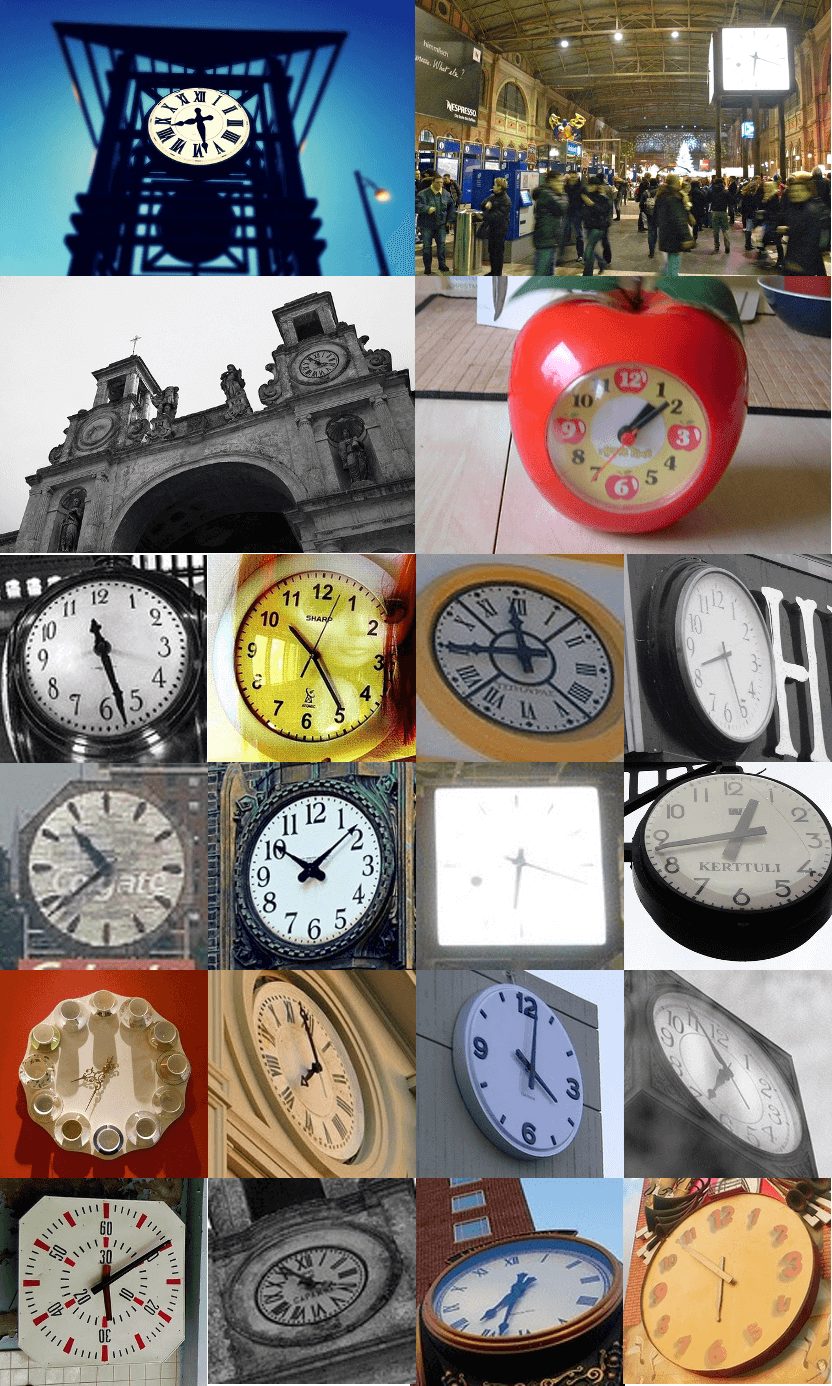}
         \caption{COCO}
     \end{subfigure}
     \hfill
     \begin{subfigure}[b]{0.33\textwidth}
         \centering
         \includegraphics[width=\textwidth]{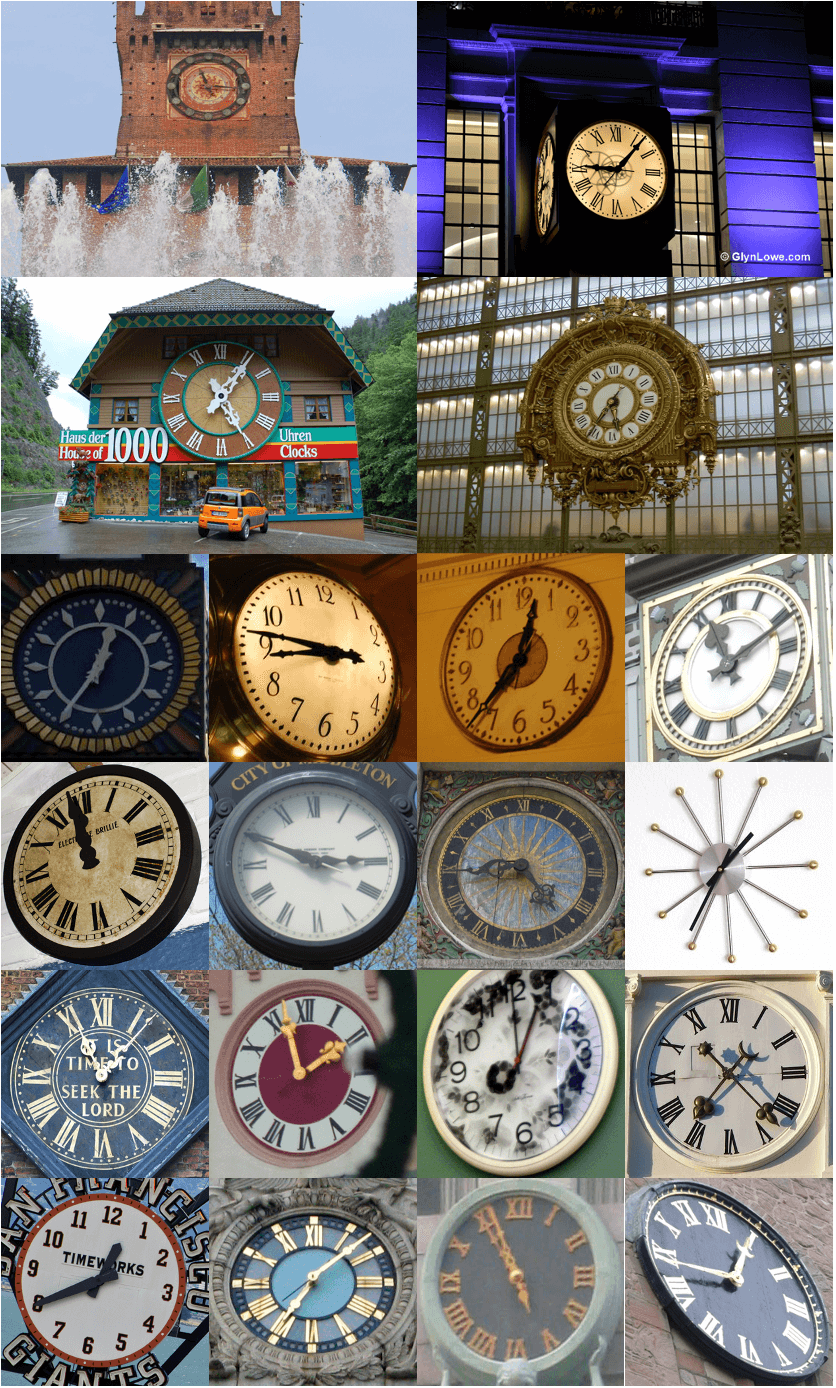}
         \caption{OpenImages}
     \end{subfigure}
     \hfill
     \begin{subfigure}[b]{0.33\textwidth}
         \centering
         \includegraphics[width=\textwidth]{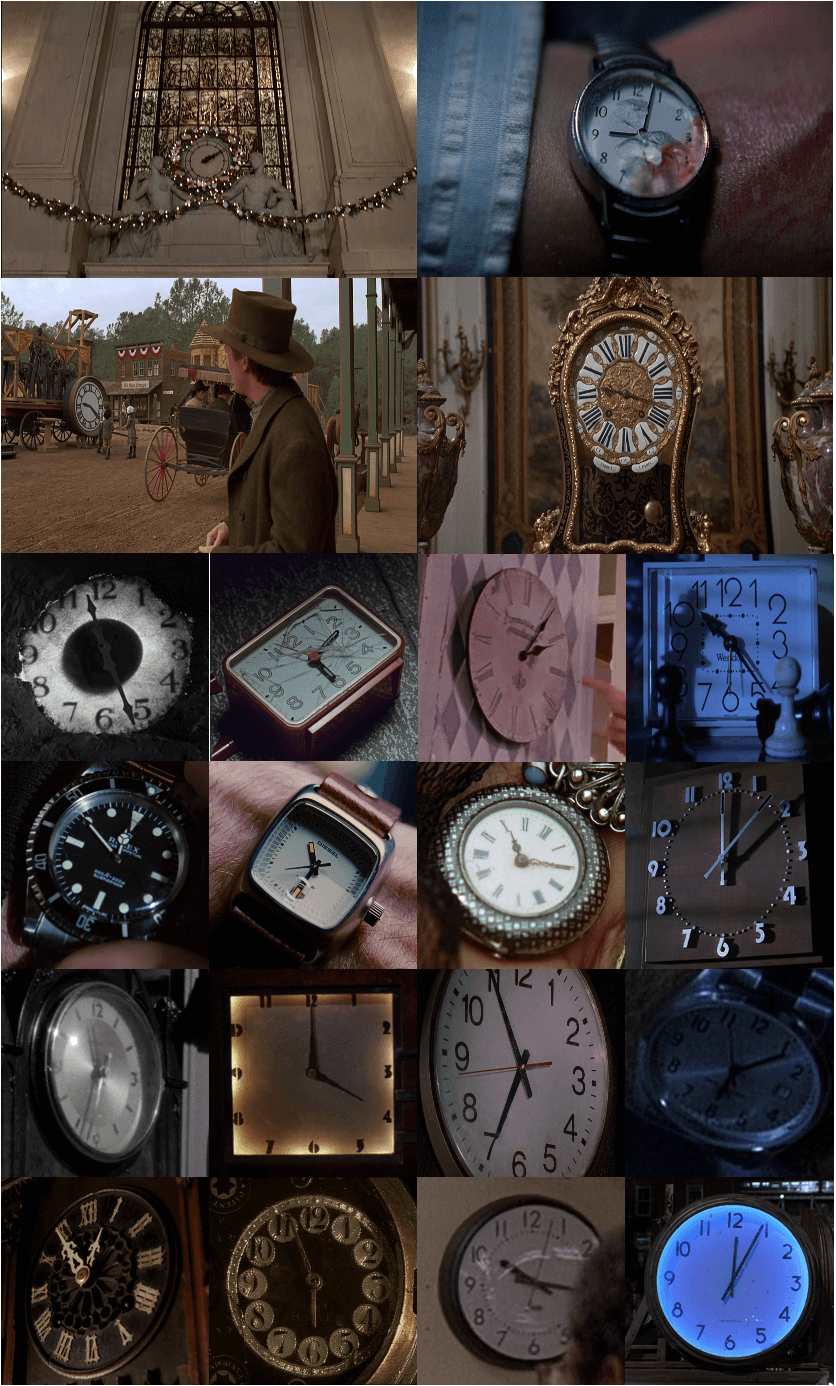}
         \caption{Clock Movies}
     \end{subfigure}
    \vspace{-.6cm}
    \caption{\small 
    \textbf{Evaluation data.} From left to right, the figure shows example scenes and cropped clocks from COCO \cite{lin2014microsoft}, OpenImages \cite{OpenImages} and Clock Movies datasets. For the first two datasets, we filter scenes with readable analog clocks from the original datasets and provide time labels, while the latter is from 600 different movies. They contain 4,472 images in total and are used for evaluation.\vspace{-.3cm}}
    \label{fig:eval_data}}
\end{figure*}

Nowadays, clocks come in two main types -- digital and analog. 
Digital clocks can be handled by text spotting methods~\cite{lyu2018mask, liao2019mask, liu2020abcnet, liu2018fots, qin2019towards} with relative ease, 
which we show in the Appendix,
but reading analog clocks is a different and challenging problem: there are significant appearance variations between clock faces (see Figures \ref{fig:synclock} and \ref{fig:eval_data}), their imaged shape and the position of the numbering is severely affected by camera viewpoint, and the presence of shadows and specular reflections add confusion with the clock hands.
While this problem has existed for a long time~\cite{github1, github2, github3}, 
no previous solutions are able to robustly read the time from clocks, apart from under extremely limited situations. 
And, somewhat surprisingly, 
reading analog clocks in unconstrained images has been largely overlooked in the computer vision literature.
Additionally, there are no reliable benchmarks for evaluation, 
hindering the research community from tackling this task.

However, there are similarities between analog clock reading and text spotting in natural scenes -- since in both cases the design (of the clock face or text font) is chosen to be readable, and in both cases there is a detection stage and then a reading stage. Clock reading has the additional challenges outlined above, but it also 
has an additional redundancy cue in that
 the position of the hour hand gives some information
about the position of the minute hand. Given the similarities between the two tasks, we start with an approach that has been successful for text spotting: using synthetic datasets \cite{gupta2016synthetic, jaderberg2014synthetic} and spatial transformer networks \cite{shi2016robust}, and ask if these ideas transfer to our task. We find that they do to an extent, and provide further contributions to bridge the Sim2Real generalisation gap. 

While being able to carry out a new task is a sufficient reward in itself, there are a number of applications that are opened up, once we are able to automate time reading in images in the wild: first, it will now be possible to offer corrections where the image's EXIF metadata differs from the time read in the image; second, in video forensics, it will now be possible to spot if the video has been tampered with if the temporal ordering does not progress monotonically 
or if there is manipulation of the speed~\cite{hosler2020detecting}; third, it provides a new method of searching, retrieving and grouping images and videos; and, finally, clocks are just a (rather difficult) instance
of an analog scale, and the methods we have proposed can be applied with a simple adaptation to other type of scales --
from scientific instruments to industrial gauges.

In this paper, we provide the first working solution to these issues. We make the following contributions:

\textit{First}, we propose a synthetic dataset generator, SynClock, that is designed to generalize to real clocks. SynClock has several controllable features that enables it to generate clocks with a wide range of designs. Moreover, we mimic difficulties faced in recognising real clocks into the generator's data augmentation process, \eg homography transformation, artefacts, shadows.

\textit{Second}, we design a two-stage framework involving detection and recognition stages. The detection can simply be an off-the-shelf object detection model. The recognition stage involves an alignment network, which is a spatial transformer network that regresses homography transformation parameters in order to make the clock fronto-parallel, and a classification network, which determines the time accurate to the minute. We show that the model is able to generalise towards real clocks with good accuracy.

\textit{Third}, we leverage the uniformity of time -- that it flows at a constant rate, in order to generate pseudo-labels on unlabelled clock videos. Specifically, we can be reasonably confident that the time labels in a video are correct if the rate of change of predicted time is constant throughout the video. To achieve this, we use a bundle adjustment algorithm to filter eligible videos and train on those with the pseudo labels. We also propose a dataset of 3,443 unlabelled clock time-lapse videos, and show that learning from pseudo-labelled real data improves the performance. We will release the raw videos, as well as reliable automatic annotations for 1.5M frames across 2,511 videos.

\textit{Fourth}, we propose three new benchmark datasets. The first two are based on existing datasets for object detection, namely COCO and OpenImages. 
We also introduce the Clock Movies dataset, based on the film The Clock (2010), which is a 24-hour montage of different movies featuring clocks\footnote{Due to copyright restrictions, we may not be able to release the frames for the Clock Movies dataset.}.
Our model achieves 80.4\%, 77.3\% and 79.0\% \mbox{top-1}  accuracy on each dataset respectively, marking the first time that analog clocks can be read successfully in unconstrained images. 

\section{Related Work}

\noindent \textbf{Analog clock reading.} 
While there exists blog posts and repositories, 
this task is still largely absent in the research literature, 
which may be because of the lack of proper benchmarks. 
Traditional methods use handcrafted methods such as edge or line detection algorithms to read clocks~\cite{github5, github6, clockreader}, 
but they only work on simple, clean, artefact-free, fronto-parallel clocks. 
Given the absence of labelled data, 
existing works usually consider to train on synthetic clocks~\cite{github1, github2, github3}, but only go as far as testing their models on synthetic test sets. These models hence are not designed with generalisation in mind, 
and usually fail to read clocks in real scenes. 
In this paper, 
in addition to adopting a more challenging data simulation pipeline,
we also leverage the uniformity property of time, 
and mix the training with data of timelapse analog clock videos.
As a result, the gap between simulation and real-world images has been largely minimised in a self-supervised manner, {\em i.e.}~without using manual annotations.\\[-6pt]

\noindent  \textbf{Dial reading.} The closet kin to our work is on automatic dial or gauge meter readings. 
These two tasks face similar challenges in overcoming the effects of blur, glare, and reflections.
The solutions proposed are somewhat similar, using neural networks \cite{Alexeev20}, projective transforms \cite{bao19}, and virtual dataset generators \cite{cai20, Howells_2021_CVPR}, which work very well for gauges with known shape and style. These have applications towards reading electricity, water or gas dials~\cite{dial1,dial2,dial3,Alexeev20}. Our task is more challenging as clock appearances vary greatly, and we further propose a method to bridge this generalisation gap.  \\[-6pt]

\noindent \textbf{Sim2Real transfer.}  
In many computer vision and robotic tasks, 
synthetic datasets prove to be a useful source for training. 
This is especially true when the ground-truth is difficult or impossible to acquire at scale, including optical flow \cite{sintel, mayer2016large}, 
and text detection~\cite{gupta2016synthetic} and recognition \cite{jaderberg2014synthetic}, pose estimation~\cite{Doersch19},
and motion segmentation~\cite{Lamdouar21}.

\begin{figure*}[t]
\centering
\includegraphics[width=\textwidth]{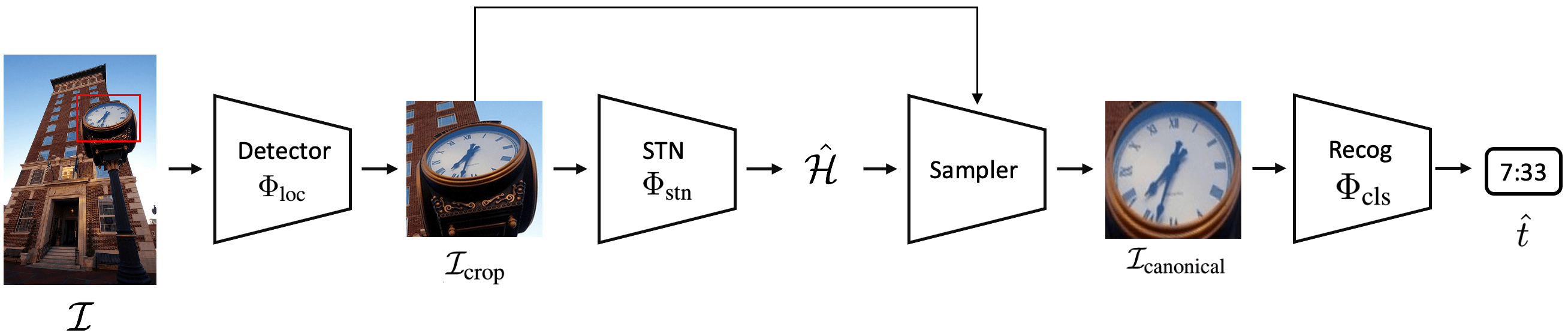}
\vspace{-.5cm}
\captionof{figure}{\small \textbf{Architecture.} Given an image $\mathcal{I}$, we first use an off-the-shelf object detector $\Phi_{\text{loc}}$ to obtain a cropped image $\mathcal{I}_{\text{crop}}$.
We then pass the cropped image to a spatial transformer network $\Phi_{\text{stn}}$ which outputs a homography matrix $\mathcal{H}$, that can then be used to warp the cropped image to a canonical view $\mathcal{I}_{\text{canonical}}$.
Lastly, the canonical image is passed to a classification network $\Phi_{\text{cls}}$ to predict the time.
\label{fig:arch}}
\end{figure*}%

\section{Architecture}
In this paper, 
our goal is to train a computer vision system that can read the time shown on the analog clock from in-the-wild images. To achieve this, we propose a framework shown in Figure \ref{fig:arch}. Specifically, our proposed architecture takes an image as input, then sequentially undergoes {\em cropping}, {\em alignment}, and {\em reading}.


\subsection{Clock Localisation Module~($\Phi_{\text{loc}}$)} 
Given an input image $\mathcal{I}$, 
we pass it through an object detection network, to localise and crop the clock. 
\begin{align}
    \mathcal{I}_{\text{crop}} = \Phi_{\text{loc}}(\mathcal{I}; \Theta_{\text{loc}}) \in \mathbb{R}^{3 \times h_{c} \times w_{c}}
\end{align}

As we will show in the experiments in Section \ref{sec:results},
the detection stage can be done using an off-the-shelf object detector \cite{liang2021cbnetv2} without much impact on performance. 
This work hence focuses on the recognition stage.


\subsection{Clock Recognition Module~($\Phi_{\text{rec}}$)}
The cropped clock could be passed directly to a classification network.
However, this is usually not ideal for two reasons, 
firstly, due to the imperfections of the localisation module; and
secondly, even if the clock is properly localised and cropped, 
it can still sometimes be difficult to read due to the viewpoint (as viewpoint alters angles). 
To overcome these problems, we adopt a spatial transformer network~(STN)~\cite{Jaderberg15} for {\em alignment}, 
facilitating the {\em recognittion}, {\em i.e.}~$\Phi_{rec}(\cdot) = \Phi_{\text{cls}}(\Phi_{\text{stn}}(\cdot))$.\\[-6pt]

\par{\noindent \bf Spatial Transformer Network~($\Phi_{\text{stn}}$)} 
takes the cropped image~($\mathcal{I}_{\text{crop}}$) as input and outputs 8 homography transformation parameters:
\begin{align}
    &\mathcal{\hat{H}} = \Phi_{\text{stn}}(\mathcal{I}_{\text{crop}}) \in \mathcal{R}^{3 \times 3} \\
    & \mathcal{I}_{\text{canonical}} = \textsc{Sampler}(\mathcal{I}_{\text{crop}}, \mathcal{\hat{H}}) \in \mathcal{R}^{3 \times h \times w}
\end{align}
where $\mathcal{\hat{H}} \in \mathcal{R}^{3 \times 3}$ refers to the predicted homography transformation with 8 degree of freedoms,
and $\textsc{Sampler}(\cdot)$ denotes a differentiable warping,
that transform the cropped clock to its canonical view~($\mathcal{I}_{\text{canonical}}$), where the clock is fronto-parallel and the `12' position is at the top. \\[-6pt]

\noindent \textbf{Classification Network~($\Phi_{\text{cls}}$).}
For reading the time,
we quantise the time and pose the recognition problem as a 720-way classification, 
{\em i.e.}~classifying both the hour~($12$) and minute~($60$) together. Specifically, 
we pass the canonical clock~($\mathcal{I}_{\text{canonical}}$) to a classification network, which outputs the probability for each class.
\begin{align}
    \hat{t} = \Phi_{\text{cls}}(I_{\text{canonical}}) \in \mathcal{R}^{720}
\end{align}
In the Appendix we compare this classification approach to time reading to a regression approach.

\noindent 
Each of the networks ($\Phi_{\text{stn}}$, $\Phi_{\text{cls}}$) is a standard ResNet-50~\cite{He16} pretrained on ImageNet~\cite{Deng09}.

After introducing the architecture, 
one question remains:
{\em how can we efficiently train this clock recognition module, 
without a laborious annotation procedure?}

\begin{figure*}[t]
{
    \centering
    \includegraphics[width=\textwidth]{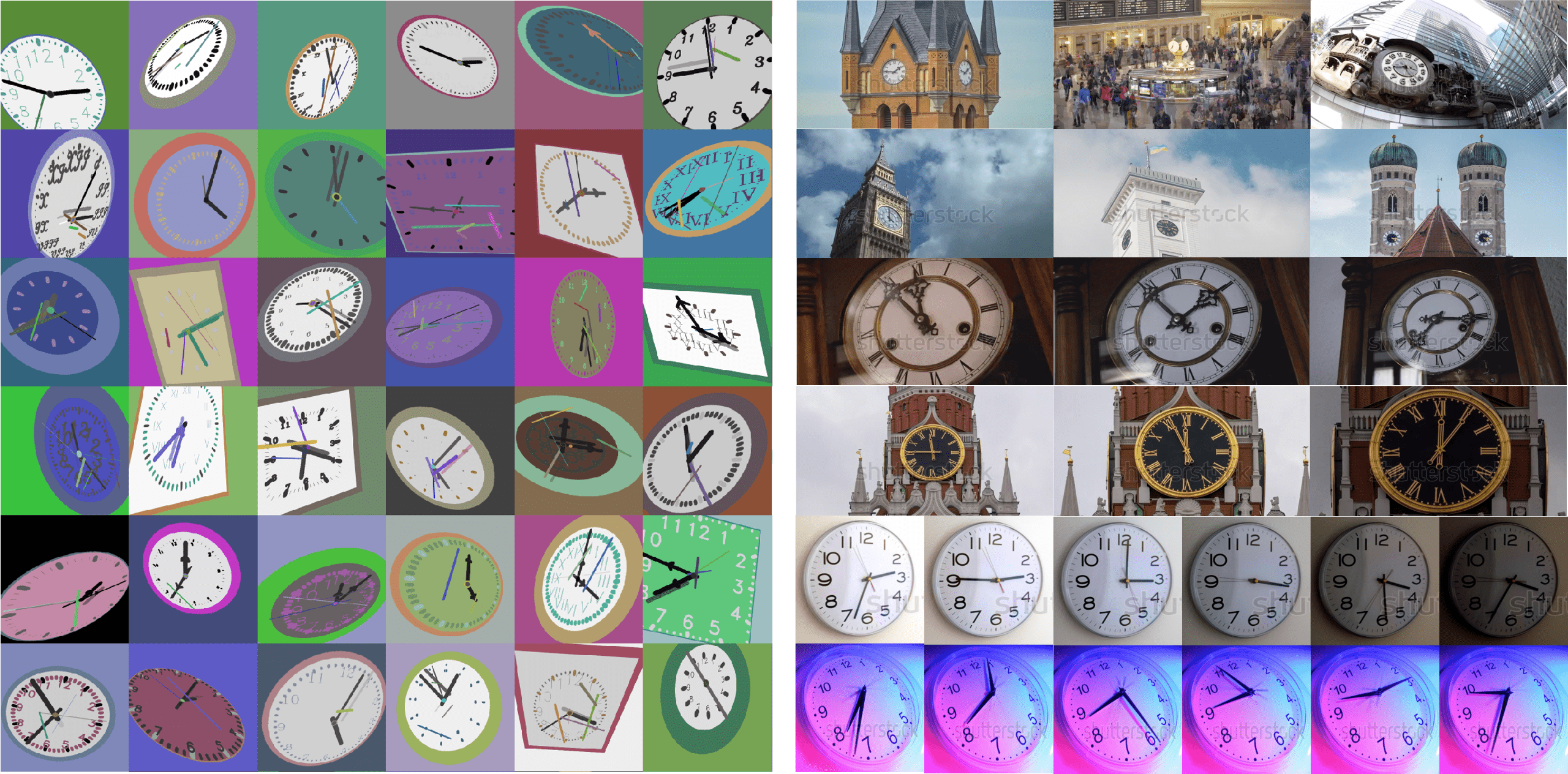}
    \captionof{figure}{\small 
    \textbf{Training data.} Left: example images from the SynClock dataset generator, which is designed to generate a wide range of clocks. We also add artefacts, such as random lines and shadows followed by augmentations. Right: example scenes from the Timelapse dataset, containing 3,443 unlabelled timelapse videos containing clocks. We train with this dataset using pseudo-labels with uniformity constraints.
    }
    \label{fig:synclock}}
\end{figure*}%

\section{Synthetic Data and Sim2Real Training}
\label{sec:sim2real}

To avoid laborious manual annotations,
we describe a procedure for training alignment and recognition with simulated clocks, advocating a Sim2Real training regime.
Specifically, 
the synthetic clocks are generated with different viewpoints, time, and styles.

\subsection{Synthetic Clock Generator~(SynClock)}
Inspired by the idea for text spotting~\cite{jaderberg2014synthetic}, 
we propose a scalable pipeline for generating synthetic analog clocks.
In order to generalise towards real clocks,
we make the dataset sufficiently diverse with several controllable parameters,
and add artefacts to simulate the real-world scenario.
Specifically, we vary these parameters during training: \\[-8pt]
\begin{itemize}[noitemsep, topsep=0pt, leftmargin=*]
\setlength\itemsep{3pt}
    \item \textbf{Background}: color. 
    \item \textbf{Clock face}: size, shape, color.
    \item \textbf{Clock border}: thickness, color.
    \item \textbf{Tick marks}: whether to have ticks every minute or every hour, gap from border, length, thickness, color.
    \item \textbf{Numerals}: whether to have numerals, gap from border, font, font size, font thickness, color.
    \item \textbf{Hands}: whether to use 2, 3, or 4 hands~(hour, minute, second, alarm), time, length, back length, thickness, color, whether to use arrow, arrow tip length, arrow size. (may be different for each hand)
    \item \textbf{Artefacts}: shadows beside a hand, random lines, 
    random homography transformation.\
    \item \textbf{Augmentation}: random blur, color channel jittering 
\end{itemize}

Figure \ref{fig:synclock} shows examples of clocks produced by this SynClock generator, and further examples are given in
the 
Appendix.

\subsection{Training on SynClock}
With full controls on the data generation procedure,
we are able to to train both spatial transformer and the classification network.
Specifically, 
$\Phi_{\text{stn}}$ and $\Phi_{\text{cls}}$ are trained with L1 loss and cross-entropy loss, 
using ground-truth transformation~($\mathcal{H}$) and time~($t$) 
generated from SynClock:
\begin{align}
\mathcal{L} = \mathcal{L}_{\text{stn}} + \mathcal{L}_{\text{cls}} 
= \sum |\mathcal{\hat{H}} - \mathcal{H}| + \sum \hat{t} \log (t)
\end{align}


\section{Pseudo-labelling Real Videos}
\label{sec:uniform}

While training on carefully designed synthetic clocks allows reasonable generalisation to real images, 
the domain gap between simulation and real images still exists,
as will be demonstrated in our experiments in Section~\ref{sec:recognition_results}.
In this section, we describe a simple idea for minimising the domain gaps by mixing the training with real video frames.
One critical issue would be,
{\em how can we obtain the ground-truth time for these video frames? }

\subsection{Uniformity Constraints}
\begin{figure*}[!htb]
\centering
\includegraphics[width=\textwidth]{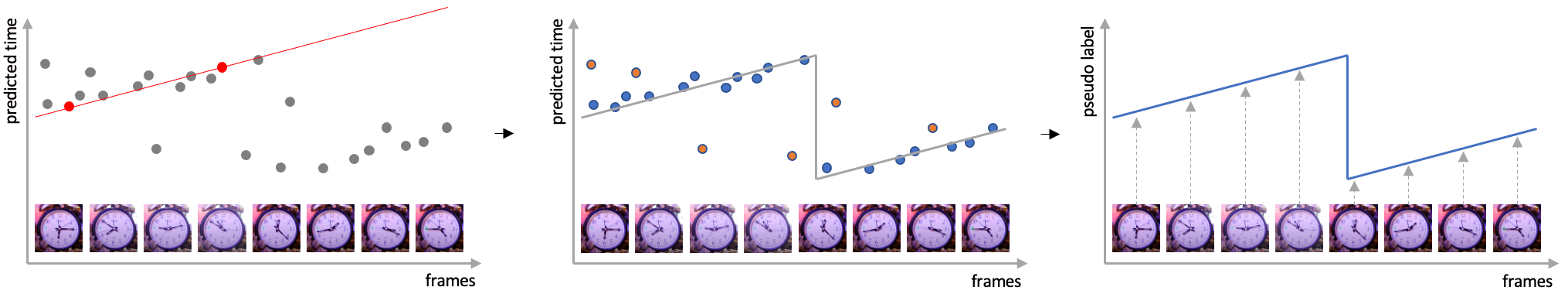}
\captionof{figure}{\small \textbf{Uniformity Constraints.} Given a timelapse video of a clock, we iteratively fit a line through randomly sampled predictions (left), before rectifying it into the valid range [0, 720) using the modulo operator and counting the inliers (middle). If the maximum inlier count is above a threshold, we then correct the readings using the fitted line (right) and add the pseudo-labelled clocks to the training set.
\label{fig:uniformity}}
\end{figure*}%
\begin{figure*}[!htb]
\centering
\includegraphics[width=\textwidth]{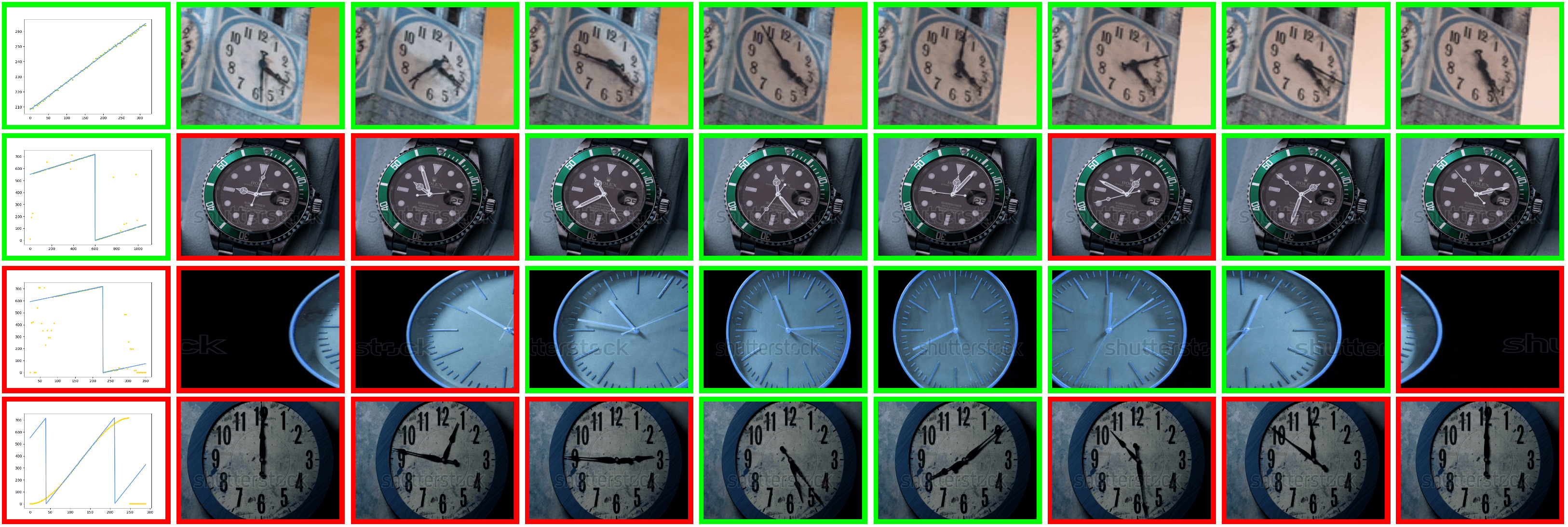}
\captionof{figure}{\small \textbf{Examples of filtering using uniformity constraints.} The first two rows show examples of videos that passed filtering, 
with incorrect predictions being calibrated accordingly. 
The bottom two rows show examples of videos that failed, 
due to out-of-frame and non-uniform speed respectively. 
The color of the box indicates success or failure. \vspace{-.3cm}}
\label{fig:filter}
\end{figure*}%

We leverage the \textbf{uniformity} property of time, 
\ie~time flows unidirectionally at a constant rate in videos.
Specifically,
given an unlabelled clock video, 
we pass each individual frame through the localisation and recognition module to get the time predictions.
As the recognition module has only been trained on synthetic data,
it is likely to generate incorrect predictions for certain frames.
Here, 
we can conveniently fit a line with RANSAC~\cite{Fischler81} -- a robust fitting algorithm that allows line fitting in the presence of outliers,  incorrect time predictions in our case. 
An iteration of RANSAC involves fitting a line with the minimal number of randomly sampled points (two), counting points within a threshold distance as `inliers', 
and the points that are distant from the line are  treated as outliers.
After running RANSAC multiple times, 
the best fitted line with maximum number of inliers are adopted, 
and all outliers can be re-calibrated accordingly, 
as shown in Figures~\ref{fig:uniformity} and \ref{fig:filter}.

Note that the data is cyclic, \ie~11:59 (719) is connected to 0:00 (0). To overcome this, we modified RANSAC such that it fits a sawtooth wave instead of a line, as seen in Figure \ref{fig:uniformity}. Specifically, after fitting a line through the randomly sampled points, if the predicted time is outside the [0, 720) range, we have to repeatedly add or subtract it by 720 so that it falls into the valid range, before counting the inliers. In practice, this can be implemented simply by applying a \textit{modulo} (\%) 720 operator on the fitted line. 

\subsection{Timelapse Dataset}
While the above method can work on any video with clocks running continuously, 
storing the videos of multiple hours long at scale would be impractical. 
Instead, we use {\em timelapse} clock videos, 
where time moves much faster, 
allowing video duration to be in the range of seconds and not hours.
We hence collect a dataset of $3,443$ unlabelled time-lapse clock videos from the internet. 
Although no absolute time information can be extracted from these unlabelled videos, knowing that the speed is constant is sufficient for our use. \\[-6pt]

\par{\noindent \bf Joint Training. }
After pseudo-labelling the videos, 
we select the ones with inlier proportion being above a threshold~(fixed at 0.7), with a tight inlier margin of +-3 minutes. We also reject videos that move too slowly, that is, less than 10 minutes throughout the video. 
We add the videos that pass filtering to the training set to re-train the model. 
This whole process is automated, 
and hence no extra manual annotation or screening is required.
Although the groundtruth transformation for aligning the clocks remains unavailable, meaning the intermediate regression loss only applies to the synthetic data, the model is still differentiable end-to-end.
\begin{align}
\mathcal{L} = \mathcal{L}_{\text{stn}} + \mathcal{L}_{\text{cls}} 
= \sum_{\text{SynClock}} |\mathcal{\hat{H}} - \mathcal{H}| 
+ \sum \hat{t} \log (t)
\end{align}

\par{\noindent \bf Iterative retraining.} The process of pseudo-labelling and retraining can be performed iteratively. Specifically, after training with real data, the model becomes more capable of reading clocks, and hence can be used to generate better pseudo-labels to train the model. This process can be iterated for further performance improvements.
\section{Experimental Setup}

\subsection{Datasets}
We use separate datasets for training and testing. We utilise one synthetic and one pseudo-labelled real dataset for training, and hence are able to train the model with \textit{zero} manual annotations.
As this is a new task, there has not been proper benchmarking in the literature. We therefore propose three different test datasets. The summary statistics are given in Table~\ref{tab:datasets}. \textbf{All} five of these datasets are contributions of this paper.

\begin{table}[!htb]
\small
\centering
\begin{tabular}{lccc|ccc}
\toprule
Dataset & Type & Size & Split &  Bbox & $\mathcal{H}$ & Time \\ \midrule
SynClock & image &	$\infty$  & train  &\xmark  & \checkmark  & \checkmark \\
Timelapse & video & 3443  & train & \xmark & \xmark & \xmark \\ \midrule
COCO & image  & 1911  & test & \checkmark  & \xmark   &  \checkmark      \\
OpenImages  & image & 	1317  & test& \checkmark   & \xmark  &  \checkmark      \\
Clock Movies & image & 	1244  & test & \xmark  & \xmark  &  \checkmark \\ \bottomrule
\end{tabular}
\vspace{-.2cm}
\caption{\textbf{Statistics of the proposed datasets.} The first two datasets are used for training, while the other three for testing.\vspace{-.5cm}}
\label{tab:datasets}
\end{table}

\subsubsection{Training}


\textbf{SynClock.} We train our recognition module using the SynClock dataset, as previously explained. We also provide ground truth homographies to train the alignment network. 

\noindent\textbf{Timelapse.} This dataset is a subset of WebVid dataset \cite{Bain21} used for video retrieval, using the ``clocks time lapse'' as the keyword for search query. It contains 3,443 unlabelled videos, and we train on this dataset using pseudo-labels, with the number of filtered videos shown in Table \ref{tab:timelapse}.

\subsubsection{Testing}

\noindent\textbf{COCO.} This dataset is a subset of COCO \cite{lin2014microsoft} dataset that contains clocks, with bounding boxes provided for 1,911 images and time manually labelled by the authors. \\[-6pt]

\noindent\textbf{OpenImages.} This dataset is a subset of OpenImages \cite{OpenImages} dataset that contains clocks, with bounding boxes provided for 1,311 images and time manually labelled by the authors.\\[-6pt]

\noindent\textbf{Clock Movies.} This dataset is based on the film The Clock (2010), which is a 24-hour film montage of different movies featuring clocks. We collect 1,244 images from over 600 different movies from the movie's Fandom page. As the timestamp within the movie reflects the absolute time by design, the time label can be implicitly obtained.

\begin{table}[]
\small
\centering
\setlength\tabcolsep{10pt}
\begin{tabular}{l|cc}
\toprule 
Iteration & Videos & Frames  \\ \midrule
Pseudo-label Round 1 & 1670 & 1.02M \\ 
Pseudo-label Round 2 & 2052 & 1.30M \\ 
Pseudo-label Round 3 & 2398 & 1.46M \\ 
Pseudo-label Round 4 & 2460 & 1.48M \\ 
Pseudo-label Round 5 & 2511 & 1.51M \\ 
 \bottomrule
\end{tabular}
\vspace{-.2cm}
\caption{\textbf{Statistics of the Timelapse dataset.} This table shows the number of videos successfully labelled in each iteration of pseudo-labelling. Each iteration uses the model of the previous row. 
\vspace{-.3cm}}
\label{tab:timelapse}
\end{table}

\subsection{Implementation details}

For the localisation stage, we use an off-the-shelf detector CBNetV2 \cite{liang2021cbnetv2} to crop the clocks, and add 20\% context to each side of the image to ensure recall. The detector is trained on COCO \cite{lin2014microsoft} and achieves the state-of-the-art performance on COCO among methods with publicly released models at the time of writing. 

We first train the recognition model on SynClock dataset. During training, the dataset is generated on-the-fly. We train the model using the Adam \cite{kingma2014adam} optimiser with learning rate 1e-4 and batch size 32 for 100k iterations.

We then use the model trained on SynClock to generate pseudo-labels for the Clock Time-lapse dataset, and filter valid ones into the training set using uniformity constraints. We then fine-tune the trained model on the enlarged training set for 20k iterations with batch size 64, with half being from SynClock and half being from the pseudo-labelled dataset. We then repeat the process where we use the newly trained model to generate pseudo-labels, and then retrain the model using the same settings as the first retraining.

\begin{figure*}[!htb]
\centering
\includegraphics[width=\linewidth]{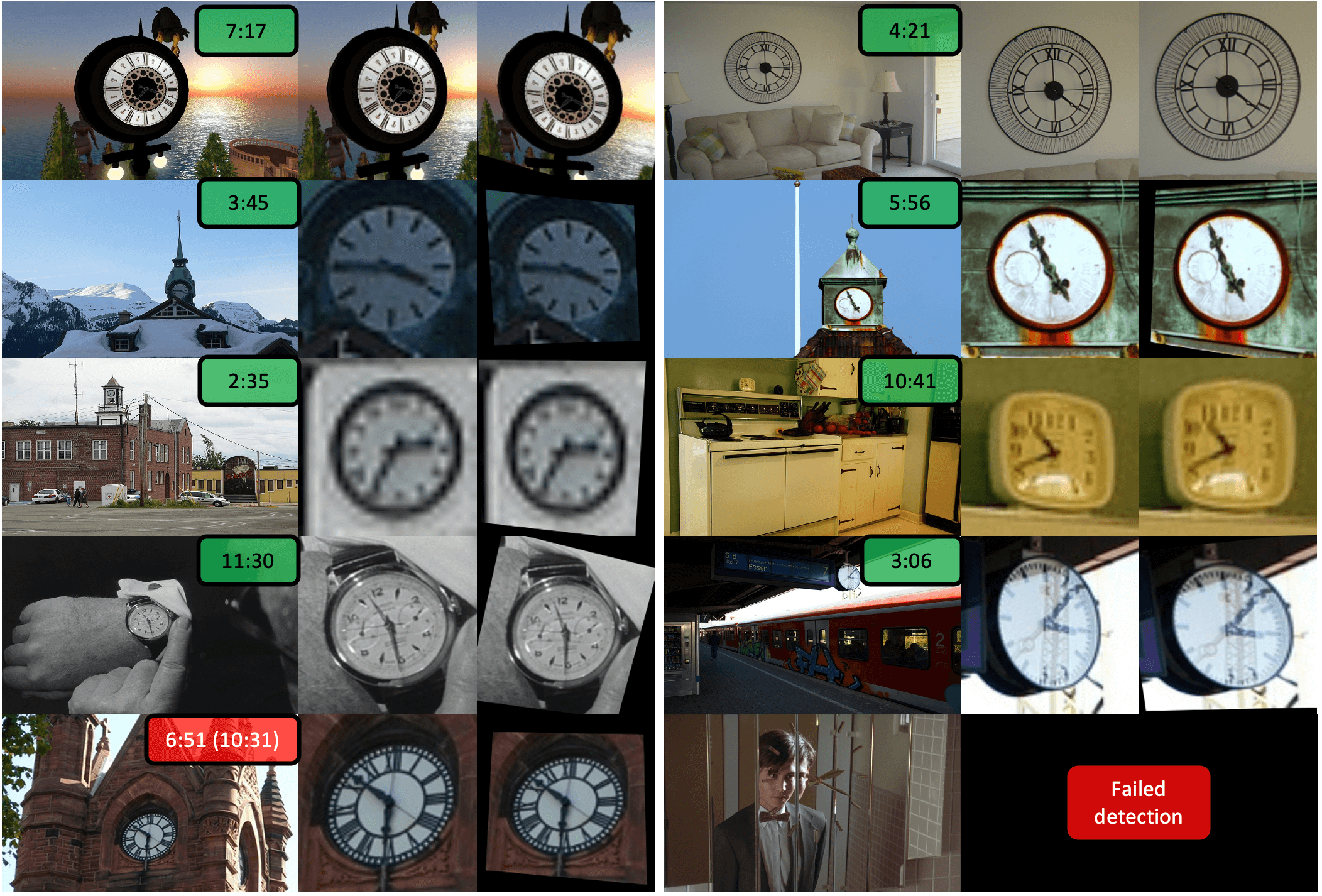}
\caption{\textbf{Qualitative results.} The columns show the original, cropped, and canonical images, with predicted time overlaid on the original image. The model is able to read clocks in varying and challenging scenes, including those with different styles (rows 1-2), low resolution (row 3), and non-frontal viewing angles (row 4). The bottom row shows two failure cases: reading swapped hands and failed detection.}
\label{fig:viz}
\end{figure*}

\subsection{Evaluation metrics}

As this is a new task, there is no existing method for evaluation. We therefore propose the following metrics:\\[-6pt]
\begin{itemize}[noitemsep, topsep=0pt]
\setlength\itemsep{2pt}
\item \textbf{Hour accuracy.} when the predicted hour is correct.
\item \textbf{Minute accuracy.} when the predicted minute is correct within a +-1 margin. We think this is close to the limit of human readability in resolving the in-between cases, especially with the absence of the second hand.
\item \textbf{Overall accuracy.} when both hour and minute are correct (minute within a +-1 margin).
\end{itemize}
We also look into top-1, 2, 3 prediction accuracies, as some clocks have ambiguities with hands.

\section{Results}
\label{sec:results}

\begin{table}[h]
\small
\centering
\setlength\tabcolsep{8pt}
\begin{tabular}{l|ccccc}
\toprule
Dataset     & 1 & 1-H & 1-M & 2 & 3 \\ \midrule
COCO      & 80.4 & 86.9 & 84.4 & 84.3 & 86.5  \\
OpenImages    &  77.3 & 83.5 & 81.9 & 81.5 & 84.6  \\
Clock Movies  &  79.0 & 82.3 & 82.4 & 83.3 & 85.5 \\ \bottomrule
\end{tabular}
\vspace{-.2cm}
\caption{\textbf{End-to-end results.} We report the accuracy on the three datasets. 1, 2, 3 are the top 1, 2, and 3 overall accuracies respectively, whereas 1-H and 1-M are the top-1 hour and minute accuracies. The model achieves success in all these datasets.}
\label{tab:main_new}
\end{table}

\begin{table}[h]
\small
\centering
\setlength\tabcolsep{13pt}
\begin{tabular}{l|ccc}
\toprule
Dataset & mAP & AP50 & AP75 \\ \midrule
COCO-val    & 77.0  & 92.1 & 74.6   \\
OpenImages & 61.5 & 89.2   &  66.4   \\ \bottomrule
\end{tabular}
\caption{\textbf{Localisation results.} 
We report the average precision (AP) of the bounding box with respect to the ground truth. Note that the object detector itself is trained on COCO-train, so only its validation set is reported. 
\vspace{-.3cm}}
\label{tab:detection}
\end{table}

\begin{table*}[ht]
\small
\centering
\setlength\tabcolsep{11pt}
\begin{tabular}[t]{l|ccccc|ccccc}
\toprule
& \multicolumn{5}{c|}{COCO~(IOU$>$50)} & \multicolumn{5}{c}{OpenImages~(IOU$>$50)}  \\ \cmidrule{1-11}
Training& 1 & 1-H & 1-M & 2 & 3& 1 & 1-H & 1-M & 2 & 3\\ \midrule
SynClock & 12.2 & 25.9 & 27.1 & 16.5 & 19.4 & 13.8 & 28.8 & 26.2 & 17.5 & 20.9 \\ 
+ Augmentation  & 16.3 & 30.7 & 29.8 & 21.5 & 25.2 & 19.0 & 33.8 & 30.9 & 23.2 & 25.9 \\ 
+ Homography  &47.8 & 58.4 & 58.6 & 57.1 & 61.9 & 44.2 & 54.6 & 56.5 & 51.5 & 56.5\\
+ Artefacts  &57.9 & 68.2 & 65.0 & 66.2 & 70.7 & 53.2 & 63.2 & 61.5 & 58.4 & 63.3   \\ 
+ Spatial Transformer & 59.6 & 71.3 & 67.0 & 67.1 & 71.4 & 55.4 & 65.8 & 64.2 & 61.8 & 65.9 \\ \midrule \midrule
+ Timelapse Round 1 \hspace{10pt} & 73.9 & 81.5 & 79.1 & 78.6 & 81.7 & 69.5 & 76.7 & 75.6 & 74.1 & 77.4\\
+ Round 2 & 80.2 & 86.3 & 84.6 & 83.9 & 86.5 & 75.8 & 81.7 & 81.3 & 80.1 & 83.1 \\ 
+ Round 3 & 82.6 & 88.1 & 85.7 & 86.2 & 88.1 & 79.0 & 83.5 & 83.5 & 81.7 & 84.5 \\ 
+ Round 4 & 82.8 & 88.4 & 86.1 & 86.4 & 88.6 & 78.9 & 83.6 & 83.5 & 82.4 & 85.5 \\ 
+ Round 5 & 82.9 & 88.4 & 86.8 & 86.6 & 88.5 & 79.6 & 84.7 & 83.9 & 83.3 & 86.2 \\ \bottomrule
\end{tabular}
\vspace{-.1cm}
\caption{\textbf{Recognition results}. 
Here, all evaluations are conducted on clocks of IOU$>$50 with respect to the groundtruth annotations, to isolate the effect of detection failure and focus solely on incorrect predictions from recognition failure.
We show that augmentation, alignment~(STN), and pseudo-labelling are all 
essential components for improving the performance of recognition module.
\vspace{-.1cm}}
\label{tab:ab1}
\end{table*}

\subsection{End-to-end Results}
\label{sec:main}

For a successful detection, both localisation and recognition have to be successful. To our best knowledge, no existing work has been able to achieve success in this setting, so it is challenging to compare to prior work. The results for the full model are shown in Table \ref{tab:main_new}. We show that the model is able to achieve success in all these challenging datasets. Note that the accuracy is lower than that reported in the last row of Table \ref{tab:ab1}, as this also includes the cases where the detection is unsuccessful.

\subsection{Localisation-only Results}
\label{sec:localisation_results}
To disentangle the effects of localisation and recognition, 
we firstly report the results for localisation on COCO and OpenImages in Table~\ref{tab:detection}. To be consistent with the object detection literature, we report the average precisions (AP50, AP75), where the bounding box IoU is over a threshold. We also report the mean average precision (mAP), which is the average of APs across the thresholds [50:5:95]. Overall, the detection task is reasonably successful.

\subsection{Recognition-only Results}
\label{sec:recognition_results}

To evaluate the performance only on recognition, we only select images where the bounding box IoU is above 50\%. 
Table~\ref{tab:ab1} shows incrementally the effects of different components within the model. \\[-6pt]

\par{\noindent \bf SynClock.}
We show the effects of parts that constitute SynClock, namely data augmentation, homography transformation, and artefacts. The homography contributes the most towards accuracy (+31.5\%/+25.2\% for COCO/OpenImages top-1 accuracy respectively) as it allows clock reading from various angles and sizes. Augmentations (+4.1\%/+5.2\%) such as blurring and jittering and artefacts (+10.1\%/+9.0\%) such as shadows and random lines both help bridge the Syn2Real generalisation gap.  \\[-6pt]

\par{\noindent \bf Spatial Transformer.}
We show the effect of the spatial transformer within the architecture, which results in a monotonic improvement across all metrics (+1.7\%/+2.2\%). \\[-6pt]

\par{\noindent \bf Pseudo-labelling Videos.}
We show that adding pseudo-labelled real video to the training set greatly contributes towards accuracy (+14.3\%/+14.1\%). 
Iteratively repeating the process also yields further improvements (+9.0\%/+10.1\%).

\subsection{Qualitative Visualisation}

\label{sec:fail}

We show the qualitative results in Figure \ref{fig:viz}, from localisation, alignment and recognition. Each process can potentially give errors, which will impact the performance. We show that our model generalises to various styles of clocks, and is able to overcome low-resolution and alignment problems. The failure case is also shown in the bottom row. 

The case where the model reads swapped hands is a good example of the limitation of our model. In this example, the hands appear to be of similar length. We (humans) can resolve this ambiguity and tell the correct time (10:31) because we look at the relative position between the hour hand and the hour mark. That is, if the time were to be 6:51 (as the model incorrectly predicted), then the hour hand should be closer to 7, and not 6. The model is not yet able to reason to this level, and hence makes wrong predictions.


\vspace{-5pt}

\section{Conclusion}
This work introduces a framework for clock reading in real images or videos. We also circumvent the lack of training data in the recognition stage by proposing the synthetic dataset generator SynClock, and iteratively pseudo-labelling on real unlabelled videos using uniformity constraints. Further, we propose three benchmark datasets with accurate time labels. In the future, clock reading should become a standard processing step for images, in the same way that text spotting and object detection are now.
\vspace{-5pt}

\paragraph{Acknowledgements.} 
We thank Yimeng Long for assistance on data annotation, Joao Carreira for an interesting discussion, 
and Guanqi Zhan, Ragav Sachdeva, K R Prajwal, and Aleksandar Shtedritski for proofreading.
This research is supported by the UK EPSRC CDT in AIMS (EP/S024050/1),
and the UK EPSRC Programme Grant Visual AI (EP/T028572/1).


{\small
\bibliographystyle{latex/ieee_fullname}
\typeout{}
\bibliography{latex/egbib}
}

\onecolumn
\section{Digital Clock Reading}
\begin{figure}[!htb]
\centering
\includegraphics[width=0.7\linewidth]{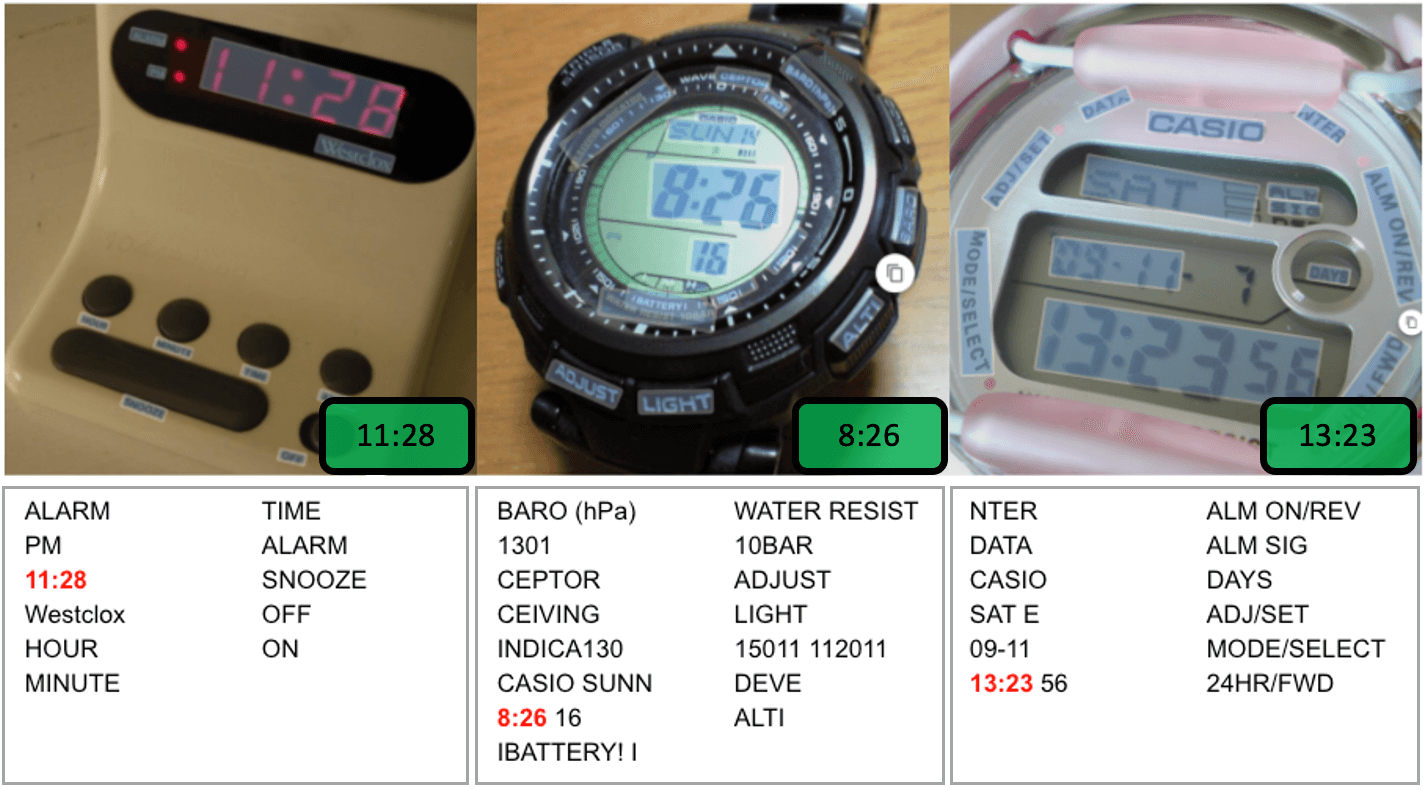}
\caption{\textbf{Reading digital clocks.} Digital clocks can be read using off-the-shelf text spotting models. The image shows examples of digital clocks, and the text extracted.}

\centering
\label{fig:digital}
\end{figure}

While this paper focuses on reading analog clocks, 
we also show that digital clocks can simply be read using off-the-shelf optical character recognition (OCR) models in Figure~\ref{fig:digital}. 
We utilise the cue that hour and minute are usually separated by a colon (:) in extracting text.

While this method allows clock reading in general, 
it is not very robust as it relies on the presence of the colon, 
which does not apply to all clocks. 
Future work can look into combining this with a detection bounding box and ensuring that the number falls into a certain range.

\section{Homography Warping}
In the alignment stage, we learn a homography matrix $\mathcal{\hat{H}}$ with 8 degrees of freedom and use that to warp the image. Projective warping with a homography is governed by the sampling equation:

{\footnotesize
\begin{equation*}
\operatorname{dst}(x, y)=\operatorname{src}\left(\frac{\mathcal{\hat{H}}_{11}^{-1} x+\mathcal{\hat{H}}_{12}^{-1} y+\mathcal{\hat{H}}_{13}^{-1}}{\mathcal{\hat{H}}_{31}^{-1} x+\mathcal{\hat{H}}_{32}^{-1} y+\mathcal{\hat{H}}_{33}^{-1}}, \frac{\mathcal{\hat{H}}_{21}^{-1} x+\mathcal{\hat{H}}_{22}^{-1} y+\mathcal{\hat{H}}_{23}^{-1}}{\mathcal{\hat{H}}_{31}^{-1} x+\mathcal{\hat{H}}_{32}^{-1} y+\mathcal{\hat{H}}_{33}^{-1}}\right)
\end{equation*}
}

One problem is that the homography matrix in the original image coordinates has a large range of values, making the regression difficult to optimise. To normalise them to a similar range, we first scale and translate the image into a [-1, 1] grid and then warp from there, before transforming back to the original space.

In implementation, this is done by transforming the predicted homography itself before warping:
$$
\mathcal{\hat{H}} = 
\begin{bmatrix}
s & 0 & st\\
0 & s & st\\
0 & 0 & 1
\end{bmatrix}
\mathcal{\hat{H}}
\begin{bmatrix}
1/s & 0 & -t\\
0 & 1/s & -t\\
0 & 0 & 1
\end{bmatrix}
$$
where the scale $s$ is equal to half the image size (224/2), and the translation $t$ is 1. This is equivalent to first scaling a 224 $\times$ 224 image to a 2$\times$2, and then translating by -1 on each axis, so that the resultant image falls into a [-1,1] grid.

\clearpage

\section{Comparison of Losses}
Intuitively, this time reading on analog clocks is more naturally one of regression rather than classification. However, empirically we find otherwise, as indicated by the comparisons in Table~\ref{tab:losses}.
In experiments A-D, we compare classification with different variants of regression losses (L1, L1 with separate hour and minute loss, L2). All the regression models fail to train on SynClock, as indicated by the low training accuracy. We then investigated a simpler case in experiments E-F, by removing augmentation, homography and artefacts while training on SynClock. The regression model~(model F) can now learn better, but still its generalization to the test data is significantly worse than the classification one.
Overall, we observe that classification is easier to train, 
and gets better performance. We think this is because regression hinders precision, as there is too little penalty for slightly wrong cases, and is hence not suitable for fine-grained recognition in our task.

\begin{table}[h]

\centering
\setlength\tabcolsep{5pt}
\begin{tabular}{c|c|c|c|ccc}
\toprule
Model & Aug. & Loss   & Train (H/M)  & 1 & 1-H & 1-M \\ \midrule
A & $\checkmark$ & cls  & 94.9/95.5   & 59.6 & 71.3  & 67.0   \\
B & $\checkmark$ & reg, L1 &  73.0/9.6 &  5.4 & 49.1 & 7.3  \\
C & $\checkmark$ & reg L2 & 60.8/6.4  & 3.2  & 30.5 & 5.8 \\
D & $\checkmark$ & reg, L1, sep & 66.7/7.5  & 4.2  & 48.2 & 6.2 \\  \midrule
E & \xmark & cls & 98.1/99.7  &  12.2 & 25.9 & 27.1 \\
F & \xmark & reg, L1 & 96.1/67.6  & 3.5  & 21.5 & 7.6 \\ \bottomrule
\end{tabular}
\vspace{-.2cm}
\caption{\textbf{Comparison between losses.} We report the train and test accuracy  for different losses. The model is trained on SynClock, and tested on COCO (IOU$>$50). }
\vspace{-.3cm}
\label{tab:losses}
\end{table}

\section{Comparison with Previous Methods}
In this section, we compare our results to a geometry-based method using Sobel edge detection and Hough transform \cite{clockreader}. 
In Table \ref{tab:compare_geometry}, we use the same detection as in our model together with the line/edge based method for recognition. 
As expected, the model struggles with variations in design, resolution, 
lighting and artefacts.

\begin{table}[h]
\centering
\setlength\tabcolsep{8pt}
\begin{tabular}{l|ccc}
\toprule
Method     & COCO & OpenImages & Clock Movies \\ \midrule
Geometry based \cite{clockreader}  &   9.7 &  7.7 & 8.2 \\
Ours & 80.0 & 77.3 & 77.4 \\ \bottomrule
\end{tabular}
\vspace{-.2cm}
\caption{\textbf{Comparison with geometry-based methods.} 
We report the Top-1 accuracy on three benchmarks.}
\vspace{-.2cm}
\label{tab:compare_geometry}
\end{table}

\section{Broader Social Impact}
As this is a new application task in computer vision, we think it is timely to also discuss the broader social impact of our work. As clocks appear in images as the background, sometimes without the subject being aware of it, it may raise privacy concerns given that it gives out more information than what is intended. Moreover, as the application progresses, there may be a possibility of combining this information with other cues such as illumination, where then it is possible to predict where and when the image is taken, raising privacy concerns.

\section{Limitations}
As introduced in the main paper, the main limitation of our model is its inability to logically reason to the level that human does, such as using the cue of relative position of hands relative to the hour mark. Our model is also unable to generalise well to clocks with unique artistic designs, where humans would understand despite not having seen one before. This includes the example in Figure \ref{fig:viz2}, where the model does not `understand' that the 'hand' of the cartoon character corresponds to the clock hand. The inability to generalise extends to unique clocks such as backward clocks, and 24-hour clocks.

\clearpage
\section{A Note on RANSAC}
\begin{figure}[!htb]
\centering
\includegraphics[width=0.4\linewidth]{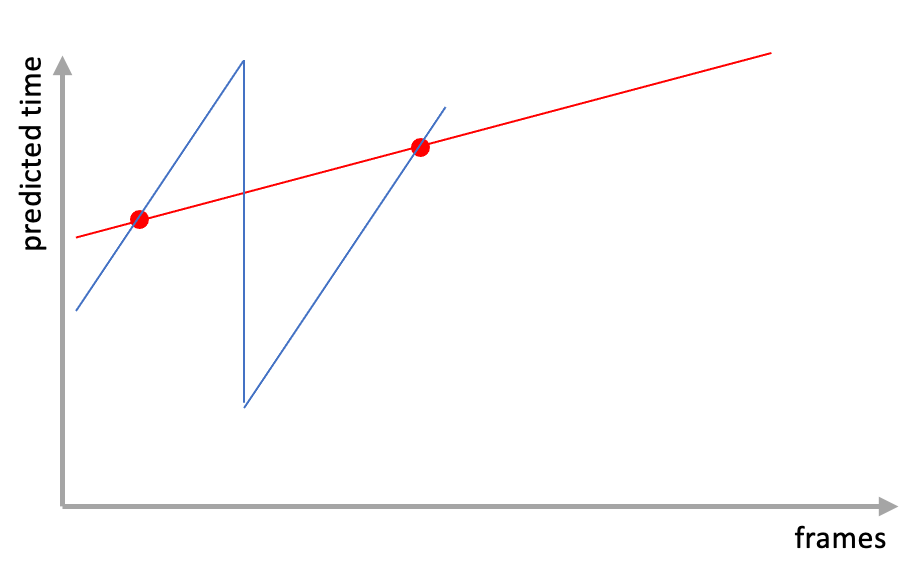}
\caption{\textbf{A note on fitting a sawtooth wave with RANSAC.} Given 2 points it is impossible to tell how the sawtooth wave will look like.}
\centering
\label{fig:ransac_note2}
\end{figure}

Normally in RANSAC, one would select the minimum number of points that fully parametrises the fit, \eg 2 for a line, 3 for a quadratic, etc. In the case of a sawtooth wave, having two points is ambiguous, as shown in 
Figure~\ref{fig:ransac_note2}, as it does not know whether the point belongs to the same period, the next one, or whether there are some periods apart, leading to infinitely many solutions. To prevent aliasing, one has to sample above the Nyquist frequency, \ie the minimum number of points must be more than 2n, where n is the number of periods.
This is highly impractical as we do not know n, and sampling that many points especially given a conservative estimate of n is very prone to errors.

\begin{figure*}[!htb]
\centering
\includegraphics[width=0.9\linewidth]{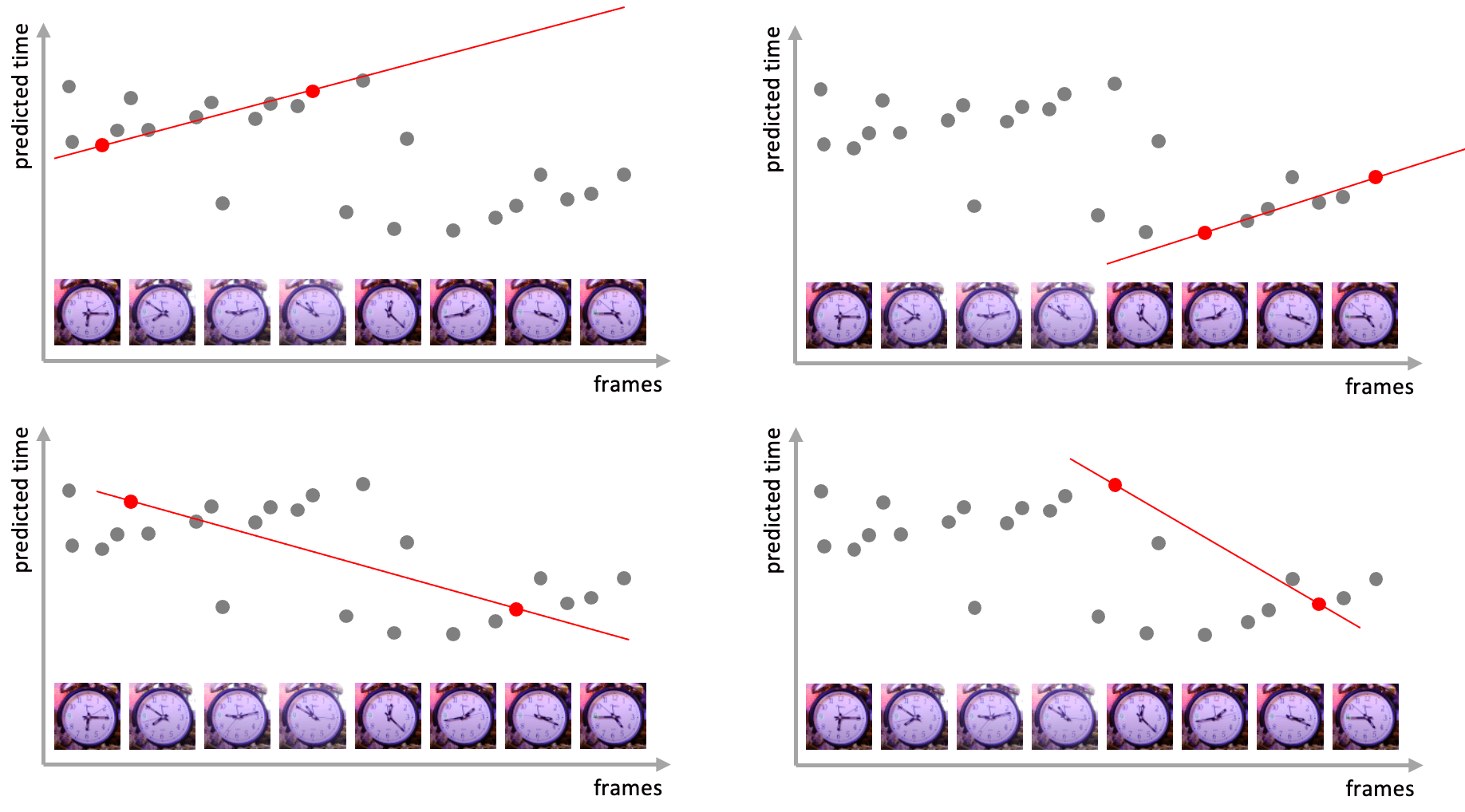}
\caption{\textbf{A note on fitting a sawtooth wave with RANSAC.} The bottom two examples are completely invalid as they violate the assumption that the points belong to the same period. Given a sufficiently large number of iterations this is not a problem.}

\centering
\label{fig:ransac_note}
\end{figure*}

Instead, we opt for a simpler solution, which is to assume that the two points sampled always belong to the same period, and apply rectification afterwards. An immediate disadvantage is that some samples are just useless, such as those in the bottom row of Figure \ref{fig:ransac_note}. However, there is approximately a 1/n chance that this assumption is valid, and this experimentally allows a good solution to be found given sufficient number of iterations, as in the top row of Figure~\ref{fig:ransac_note}.

While there are other heuristics that can be incorporated, such as sampling neighbouring points more frequently or assuming that time always flows forwards, we find that this simple solution is elegant and sufficient for our use.

\newpage

\section{More examples of RANSAC fitting}
\begin{figure*}[!htb]
\centering
\includegraphics[width=1.0\linewidth]{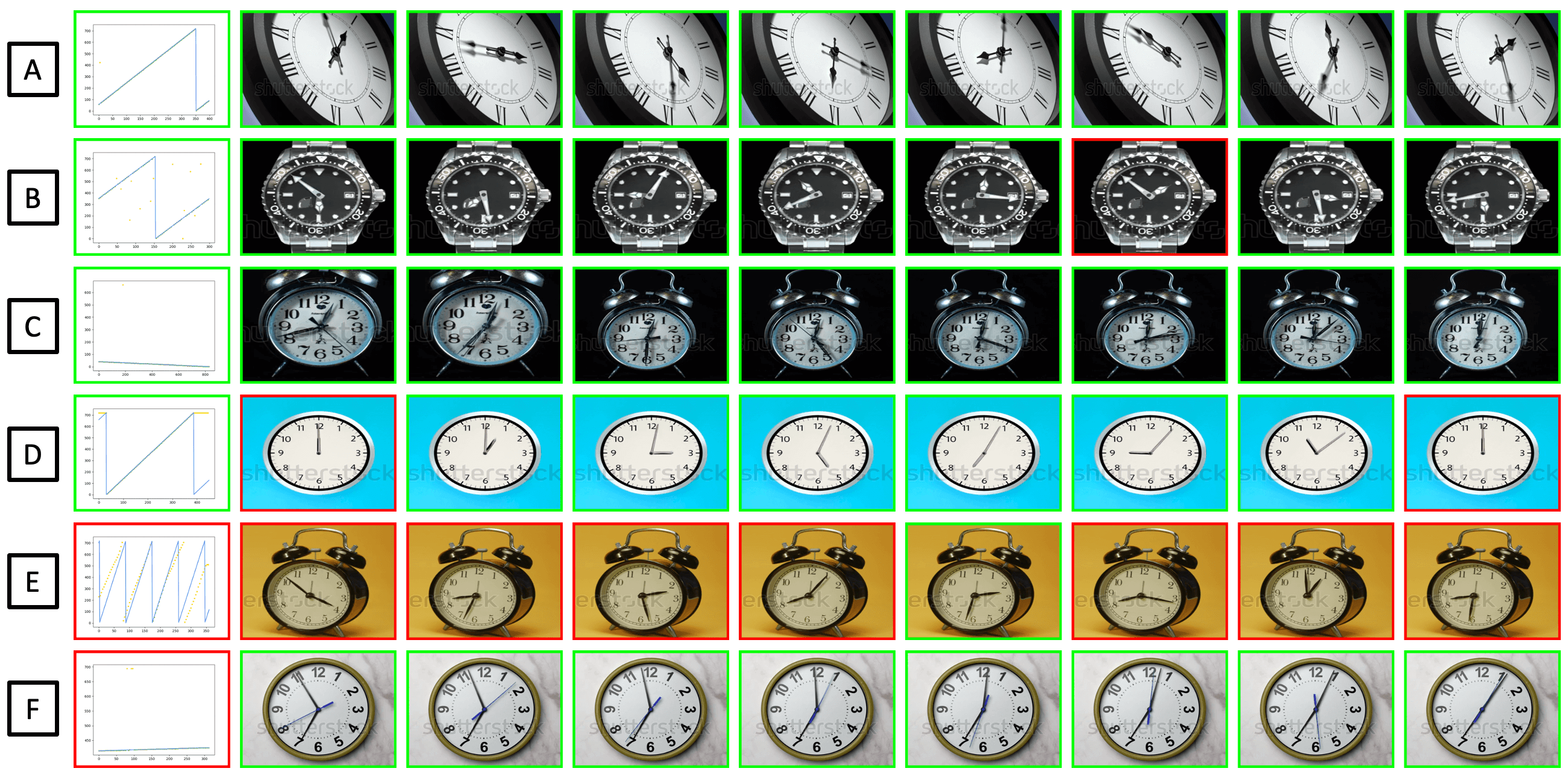}
\caption{\textbf{More RANSAC examples.} Rows A, B and C show examples of correctly filtered videos. Interestingly, the clock in row C flows backwards, but since each individual frame still reads the time correctly, the video is still correctly accepted. Row D shows a false positive case, where the clock is static at the beginning and the end of the video, but is still wrongly accepted to the dataset as the inlier proportion is above the threshold. Row E shows an example where the clock moves uniformly, but RANSAC fails to fit a line within the iteration limit. Therefore the clock is not included in the pseudo-labelled training set even though it qualifies. Row F shows a clock where we exclude from the training set because time moves too slowly, giving not many variations in time, despite being able to fit a line through.}

\centering
\label{fig:ransac2}
\end{figure*}

Figure \ref{fig:ransac2} shows more examples of video filtering using uniformity constraints. While most videos are correctly filtered, there is some uncommon false positive cases, such as when the clock stops or moves at a different speed towards the start or the end. Accepting part of such videos will be an interesting direction for future work.

\clearpage

\section{More SynClock Examples}
\begin{figure*}[!htb]
\centering
\includegraphics[width=\linewidth]{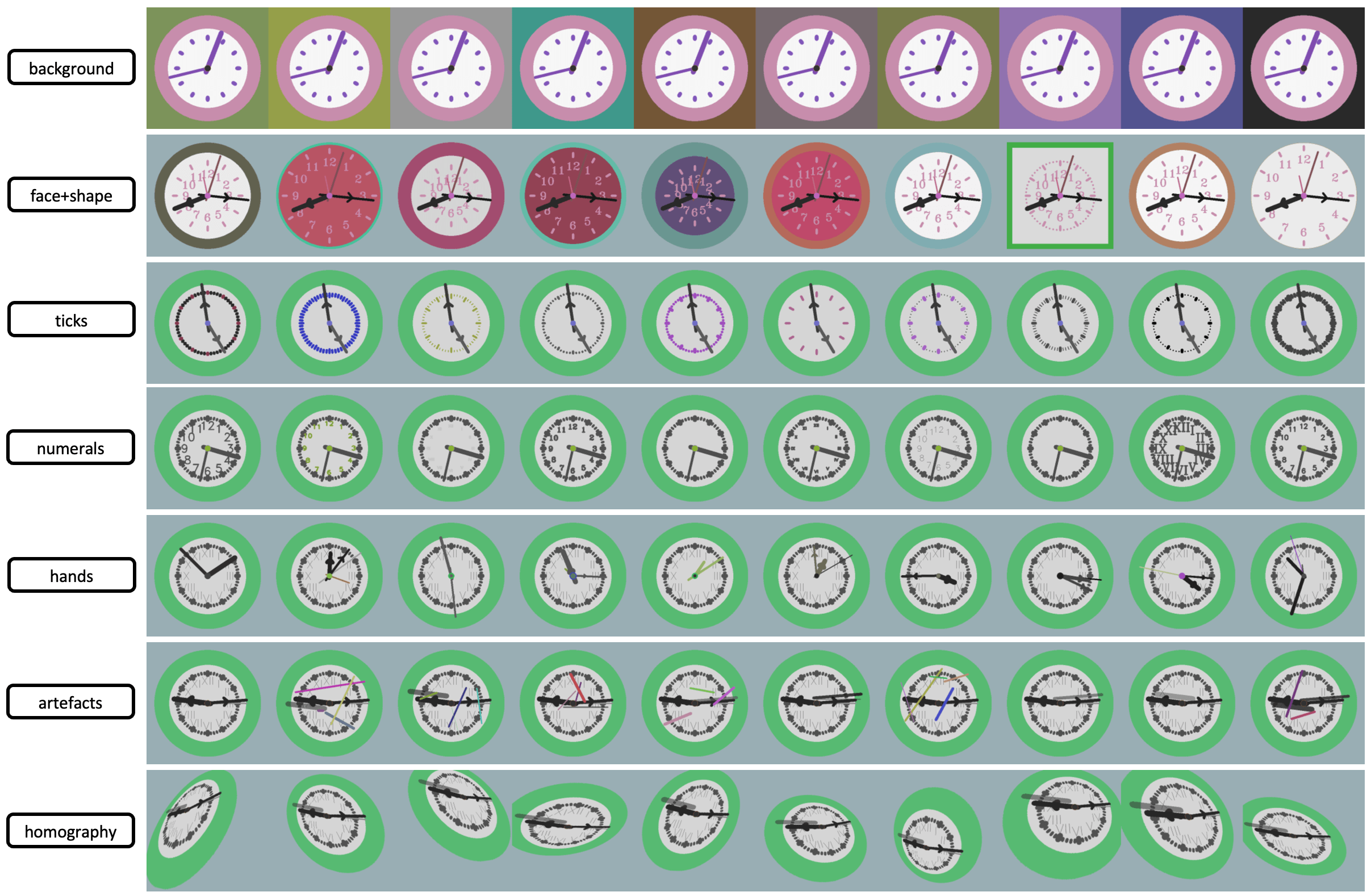}
\caption{\textbf{SynClock examples.} From top to bottom row, we show variations in background colour, clock face and shape, clock ticks, numerals, clock hands, artefacts, and homography warping.}
\label{fig:synclock2}
\end{figure*}

We show more examples of images generated from SynClock in Figure \ref{fig:synclock2}, varying one parameter at a time.

\clearpage

\section{More Qualitative Results}
\begin{figure*}[!htb]
\includegraphics[width=\linewidth]{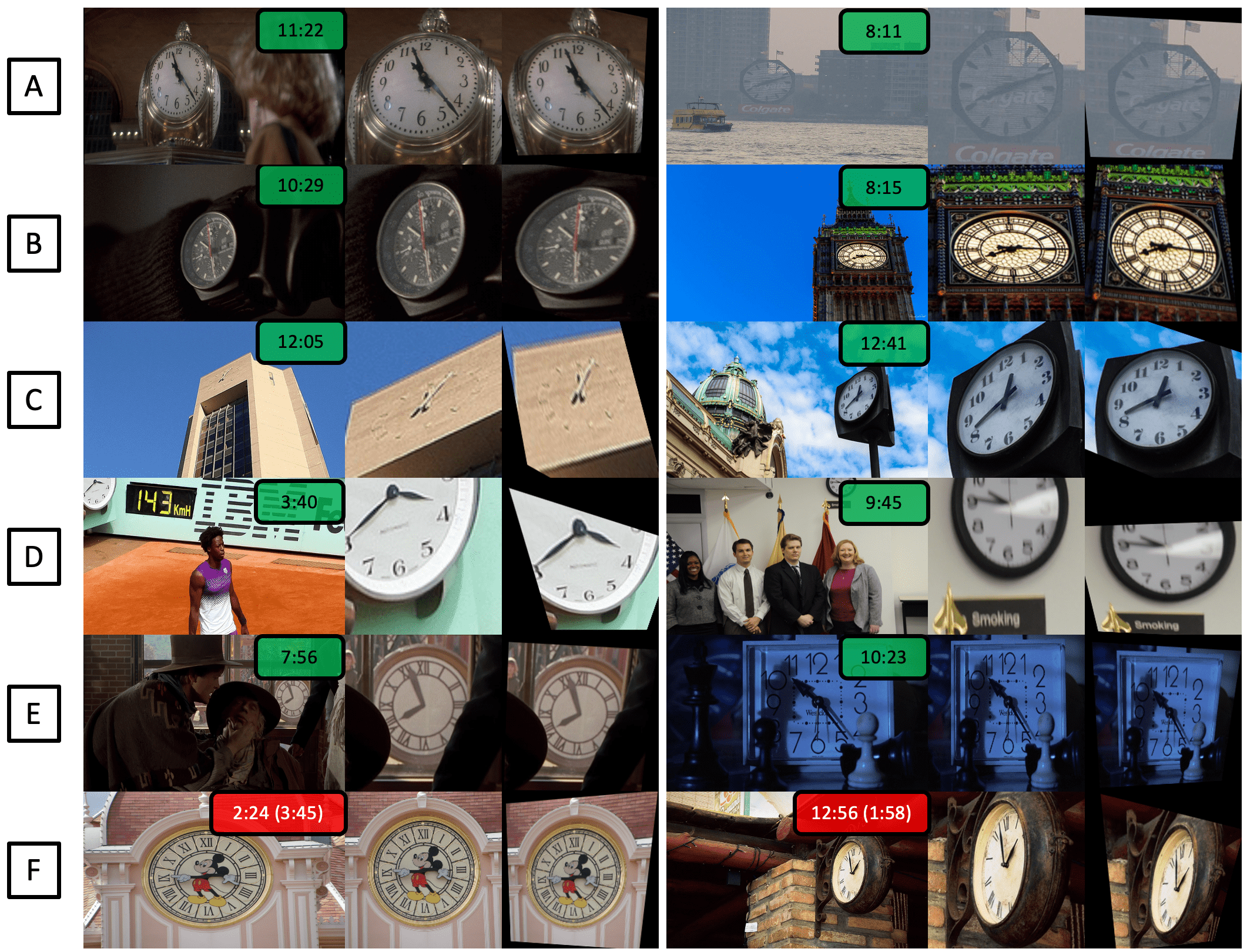}
\caption{\textbf{More Qualitative Results.}  The columns show the original image, the cropped image with 20\% context, and the canonical image. The predicted time reading is overlaid on the original image. We show that the model is able to robustly read clocks with different styles (row A), angles (rows B-C) and occlusion (rows D-E). In the last row (row F), we show a clock style that the model fails to read (left), and a case of too extreme angle (right).}
\label{fig:viz2}
\end{figure*}

We show more qualitative results in Figure \ref{fig:viz2}. 

\clearpage

\end{document}